\documentclass[12pt]{iopart}
\usepackage{graphicx}
\usepackage{eqnarray}
\usepackage{booktabs} 
\usepackage{tabularx}   
\usepackage{amsmath}
\usepackage{todonotes}
\usepackage{url} 
\usepackage{hyperref} 
\newcommand{\gguide}

\begin{document}

\title[Evaluating Visual Mathematics Problems with MLLMs]{Evaluating Visual Mathematics in Multimodal LLMs: A Multilingual Benchmark Based on the Kangaroo Tests}

\author{A. Igualde-Sáez$^1$, L. Rhomrasi$^2$, Y. Ahsini$^2$,
R. Vinuesa$^3$, S Hoyas$^2$, J.P. García-Sabater$^1$, M.J. Fullana-i-Alfonso$^4$ and J.A. Conejero$^2$}

\address{$^1$ Grupo de Investigación en Reingeniería, Organización, trabajo en Grupo y Logística Empresarial - ROGLE, Universitat Polit\`ecnica de Val\`encia, 46022, Valencia (Spain)}
\address{$^2$ Instituto Universitario de Matemática Pura y Aplicada, Universitat Polit\`ecnica de Val\`encia, 46022, Valencia (Spain)}
\address{$^3$ FLOW, Engineering Mechanics, KTH Royal Institute of Technology, SE-100 44, Stockholm, Sweden}
\address{$^4$ Institut Universitari de Matemàtica Multidisciplinària, Universitat Polit\`ecnica de Val\`encia, 46022, Valencia (Spain)}
\ead{aconejero@upv.es}

\begin{abstract}
Multimodal Large Language Models (MLLMs) promise advanced vision-language capabilities, yet their effectiveness in visually presented mathematics remains underexplored. This paper analyzes the development and evaluation of MLLMs for mathematical problem-solving, focusing on diagrams, multilingual text, and symbolic notation. We then assess several models—including GPT-4o, Pixtral, Qwen-VL, Llama 3.2 Vision variants, and Gemini 2.0 Flash—in a multilingual Kangaroo-style benchmark spanning English, French, Spanish, and Catalan. Our experiments reveal four key findings. First, overall precision remains moderate across geometry, visual algebra, logic, patterns, and combinatorics: no single model excels in every topic. Second, while most models see improved accuracy with questions that do not have images, the gain is often limited; performance for some remains nearly unchanged without visual input, indicating underutilization of diagrammatic information. Third, substantial variation exists across languages and difficulty levels: models frequently handle easier items but struggle with advanced geometry and combinatorial reasoning. Notably, Gemini 2.0 Flash achieves the highest precision on image-based tasks, followed by Qwen-VL 2.5 72B and GPT-4o, though none approach human-level performance. Fourth, a complementary analysis aimed at distinguishing whether models reason or simply recite reveals that Gemini and GPT-4o stand out for their structured reasoning and consistent accuracy. In contrast, Pixtral and Llama exhibit less consistent reasoning, often defaulting to heuristics or randomness when unable to align their outputs with the given answer options.
\end{abstract}
\vspace{10pt}

%

\section{Introduction}

Recent advances in Multimodal Large Language Models (MLLMs) have demonstrated remarkable capabilities across a wide range of tasks involving both text and images~\cite{zhang2024mmllms}. Models such as GPT-4V~\cite{achiam2024}, Qwen-VL~\cite{qwenvl2.5-7b}, and LLaVA Vision~\cite{liu2023llava} have shown promising results in bridging high-level linguistic reasoning with visual understanding. This reasoning ability involves pattern recognition, abstract thinking, and the application of mathematical and logical rules. In the case of Math Olympiad problems, solutions typically require not only advanced mathematical knowledge but also a significant degree of creativity~\cite{andreescu2008mathematical}. Therefore, mathematical problems are crucial for evaluating the logical reasoning capabilities of MLLMs \cite{li2024survey}.\medskip

The extent to which these models truly understand advanced mathematical problems remains largely unexplored, particularly in scenarios involving complex diagrams, formulas, and mathematical notation. While text-based Large Language Models (LLMs) have demonstrated progress in solving such problems using techniques like chain-of-thought prompting and model fine-tuning~\cite{peng2024multimath,liu2025arithmetic}, their effectiveness tends to decline significantly when visual components are introduced.\medskip

Interpreting geometric diagrams, axis-plotted integrals, or combinatorial grids presents a substantial challenge for multimodal models, which must effectively capture and integrate both textual and visual information~\cite{shi2024math-llava,shindo2024diff}. In many cases, these models fail to incorporate diagrammatic content into their reasoning processes. As a result, their performance on problems that include images often remains similar to their results on text-only versions, suggesting that they either overlook or are unable to use crucial visual cues~\cite{gao2023gllava}.\medskip

Numerous mathematical benchmarks have been developed to assess the performance of LLMs; however, in this work, we focus on a benchmark traditionally used to measure human reasoning skills: the Kangaroo Mathematics Competition (KMC)~\cite{KMC}. Widely recognized for testing the logical reasoning abilities of secondary school students, the KMC presents problems that frequently require interpreting embedded figures, which are often essential for arriving at the correct solution. Further, the KMC problems are not divulged before the competition, for obvious reasons, so they are an escellent text to check for recitation~\cite{yan2025recitationreasoningcuttingedgelanguage}.\medskip

We have recently investigated the ability of LLMs augmented with dedicated reasoners to solve text-only Kangaroo benchmark problems and found that reasoners significantly outperformed their non-reasoner counterparts~\cite{rhomrasi2025llm}. In fact, some of the non-reasoner models obtained the same score as a random-guessing student.\medskip

Drawing on these insights, the present study extends the evaluation to the multimodal case, incorporating problems that contain figures or diagrams within images from the Kangaroo contests. Our analysis covers both open-source models (Meta, Llama, Alibaba Qwen, Mistral Pixtral) and closed-source offerings (OpenAI GPT, Google Gemini). All these methods are described below. In addition to the methods, we have tested languages with both global and regional significance, such as English, Spanish, French, and Catalan.\medskip

The results highlight a broader trend: while state-of-the-art models such as Gemini 2.0 have higher accuracy, they still struggle when the problem requires substantial interplay between image perception and step-by-step mathematical reasoning.
Besides, our comparison analyzes the performance of open and closed-source models under text-only and text+figure problems. We expect that our findings permit us to illustrate existing models' current strengths and limitations when integrating visual and textual information for reasoning.\medskip

This work is structured as follows: In Section \ref{sec:mlmm}, we revise the latest advances in LLMs' visual, mathematical reasoning. Section \ref{sec:methodology} outlines our evaluation approach using multilingual Kangaroo mathematics datasets. Section \ref{sec:results} presents detailed analyses comparing model performance across modalities and topics. Finally, Section \ref{sec:discussions} addresses implications and limitations and suggests future research directions.\medskip

\section{State of the art: Multimodal Large Language Models for Mathematical Reasoning}
\label{sec:mlmm}

Mathematical reasoners designed to tackle mathematical problems often included, as part of their pipeline, a computer vision module for extracting geometric or symbolic representations, which were subsequently passed to a solver to generate a proof-like answer~\cite{Seo2015,Sachan2017}.
 Pioneering multimodal systems, such as Flamingo~\cite{alayrac2022flamingo}, integrated high-capacity text-only models with dedicated visual encoders to perform tasks like image captioning and Visual Question Answering (VQA)~\cite{antol2015vqa}.\medskip

Several other mathematical benchmarks have also been proposed. For instance, Geometry3K~\cite{lu2021inter} enables the evaluation of geometry-solving skills. Lu et al.~introduced MathVista~\cite{lu2023mathvista}, a dataset that encompasses a broad range of mathematical tasks, with a particular emphasis on visual understanding rather than pure mathematical reasoning. Recent studies have begun to investigate the extent to which MLLMs can comprehend visual diagrams in the context of mathematical reasoning, where problems often feature complex diagrams and formulas. Although many models excel at reading text within images, they frequently struggle to interpret geometric diagrams and mathematical symbols when multi-step reasoning is required~\cite{zhang2024mathverse}. Fan et al.~\cite{fan2024nphardeval4v} introduced NPHardEval4V, which integrates algorithmic problems with visual representations to assess reasoning capabilities. Additionally, Zhou et al.~\cite{zhou2024your} proposed a comprehensive checklist for evaluating the robustness of reasoning across different models.\medskip

Despite rapid gains on standard benchmarks (e.g., VQAv2 \cite{goyal2017making}, COCO Captions \cite{chen2015microsoft}, ScienceQA \cite{saikh2022scienceqa}), deeper evaluations indicate that many models still default to superficial pattern matching rather than genuinely reasoning about visual content and math tasks ~\cite{caffagni2024the}, such as diagrams in geometry, charts in algebraic contexts, and multilingual notations require robust cross-modal alignment~\cite{Lu2022scienceQA}. More recent neural approaches prefer end-to-end frameworks yet still show limited effectiveness.\medskip

Recently, the emergence of LLMs has propelled research in domains where language is only one component of the signal~\cite{gemini2024}.
The aforementioned MLLMs--- LLaVA, MiniGPT-4, Qwen-VL---- leveraged pretrained foundation models like Llama to create powerful vision-language assistants capable of following open-ended instructions. G-LLaVA~\cite{gao2023gllava} and Math-LLaVA~\cite{shi2024math-llava} adapt LLaVA-based architectures for geometry and math, respectively, but their performance lags behind dedicated reasoners when diagrams and images are crucial to the solution. A notable gap remains between text-only mathematics performance (which can be quite high) and truly multimodal math performance, where diagram precision is paramount~\cite{Trinh2024}.\medskip

Additional complexity arises with multilingual or non-Latin characters in images. While some models (e.g., Qwen-VL) focus on OCR-like capabilities for multiple languages~\cite{bai2023qwen-vl}, others either default to English or fail to read local notation accurately~\cite{peng2024multimath, zhang2024mathverse}. This shortfall becomes particularly problematic in problem sets like KMC, which deliver visually embedded questions with mathematical notation and are available in different languages. Successful reasoning thus requires a combination of multilingual text interpretation, geometry or symbolic parsing, and an ability to integrate these elements in step-by-step logical deductions.\medskip

We have selected a combination of open-source models and cutting-edge multimodal LLMs to compare their performance, see \autoref{table:tab1} and references inside. The evaluated models ranged from compact variants, such as Pixtral-12B and Qwen-VL 2.5 7B, to mid-sized systems, including Llama 3.2 Vision (11B and 90B) and Pixtral-Large, as well as larger state-of-the-art models, specifically GPT-4o, Gemini 2.0 Flash (including the Lite variant), and Qwen-VL 2.5 72B.\medskip

\begin{table}[t]
\centering
\caption{
Overview of the evaluated AI models. The table includes compact models such as Pixtral-12B and Qwen-VL 2.5 7B, mid-sized systems like LLaMA 3.2 Vision (11B and 90B) and Pixtral-Large, and larger state-of-the-art multimodal architectures, including GPT-4o, Gemini 2.0 Flash (and its Lite variant), and Qwen-VL 2.5 72B. For each model, the number of parameters, developing company, and references are provided. The last column presents the results of every method in the KMC; the left side of this column corresponds to image-based questions, while the right side corresponds to non-image questions.
}
\label{table:tab1}

\begin{tabular}{lllll}
\toprule
\textbf{Model} & \textbf{Parameters} & \textbf{Company} & \textbf{Reference} & \textbf{Results (\%)} \\
\midrule
Pixtral-12B & 12B & Mistral AI & \cite{pixtral12b} & 24.5 / 27.8\\
Qwen-VL 2.5 7B & 7B & Alibaba (QwenLM) & \cite{qwenvl2.5-7b} & 32.5 / 47.3\\
LLaMA 3.2 Vision 11B & 11B & Meta AI & \cite{llama3.2-11b} & 14.0 / 15.8\\
LLaMA 3.2 Vision 90B & 90B & Meta AI & \cite{llama3.2-90b} & 15.0 / 16.5\\
Pixtral-Large & 48B (estimated) & Mistral AI & \cite{pixtral-large} & 29.2 / 39.3\\
GPT-4o & Unknown & OpenAI & \cite{gpt4o} & 40.2 / 65.3\\
Gemini 2.0 Flash & Unknown & Google DeepMind & \cite{gemini2.0-flash} & 45.4 / 75.9 \\
Gemini 2.0 Flash Lite & Unknown & Google DeepMind & \cite{gemini2.0-flash-lite} &39.4 / 66.5\\
Qwen-VL 2.5 72B & 72B & Alibaba (QwenLM) & \cite{qwenvl2.5-72b} & 43.5 / 70.6\\
\bottomrule
\end{tabular}
\end{table}

\section{Methodology and Evaluation}
\label{sec:methodology}

Initially inspired by the Australian Mathematics Competition \cite{AMC}, the Kangaroo tests emphasize logical reasoning skills over routine calculations, making it highly suitable for evaluating the reasoning capacities of LLMs. The competition consists of 30 multiple-choice questions arranged in order of increasing difficulty, each requiring participants to choose the correct answer from several provided options. Table~\ref{tab:lang_levels} summarises the names of the different KMC tests conducted, along with their corresponding age-level equivalences.\medskip

\begin{table}
\caption{\label{tab:lang_levels}Comparison of language levels and age ranges across English, French, Spanish, and Catalan curricula.}
\centering
\begin{tabular}{@{}ll ll ll ll}
\br
    \centre{2}{\textbf{English}} & \centre{2}{\textbf{French}} & \centre{2}{\textbf{Spanish}} & \centre{2}{\textbf{Catalan}} \\
\br
\textbf{Level} & \textbf{Age} & \textbf{Level} & \textbf{Age} & \textbf{Level} & \textbf{Age} & \textbf{Level} & \textbf{Age} \\
\mr
Felix       & 6--8 yrs     & Écoliers    & 10--11 yrs   & Nivel 1 & 12--13 yrs  & Nivell 1 & 12--13 yrs \\
Ecolier     & 8--10 yrs    & Benjamins   & 11--13 yrs   & Nivel 2 & 13--14 yrs  & Nivell 2 & 13--14 yrs \\
Benjamin    & 10--12 yrs   & Cadets      & 13--15 yrs   & Nivel 3 & 14--15 yrs  & Nivell 3 & 14--15 yrs \\
Kadett      & 12--14 yrs   & Juniors     & 15--17 yrs   & Nivel 4 & 15--16 yrs  & Nivell 4 & 15--16 yrs \\
Junior      & 14--16 yrs   & Étudiants   & 17--18 yrs   & Nivel 5 & 16--17 yrs  & Nivell 5 & 16--17 yrs \\
Student     & 16--18 yrs   &             &              & Nivel 6 & 17--18 yrs  & Nivell 6 & 17--18 yrs \\
\br
\end{tabular}
\end{table}

A dataset comprising KMC tests conducted in English, Spanish, French, and Catalan from 2014-2024 was compiled and structured for this study. These tests correspond to those administered in Australia  \cite{EMC}, Spain \cite{SMC}, France \cite{FMC}, and the Comunitat Valenciana \cite{VMC}, respectively. The dataset consists of a CSV file, where each row includes the language, the original question, the multiple-choice options, the ground truth, the assigned score, grade level, and an indication of whether figures or images are involved. Questions containing figures or images, including those within answer choices, were collected separately in a complementary repository of images organized according to each language. Additionally, image-based questions were systematically categorized according to their mathematical content and reasoning demands. The distinctions between these categories, including examples and descriptions, can be observed in Table~\ref{tab:mathematical_benchmarks}.\medskip

In assessing mathematical competencies, specific visual elements align with distinct subject areas. Geometry and Figures questions typically feature squares, octagons, and three-dimensional objects, culminating in problems requiring perimeter calculations and geometric element counting. Visual Algebra and Arithmetic questions incorporate planar figures such as triangles and rectangles, along with segments, midpoints, and angles. This leads to final problems where students must determine unknown measurements, including lengths, areas, and angle calculations. Visual Logic and Reasoning questions commonly present images depicting people, arrows, and symmetrical figures, asking students to identify relationships or patterns in visual information. Patterns and Sequences questions feature figures such as arrows, rhombuses, or circles, challenging students to identify the underlying pattern and determine the final value in a sequence. Finally, Combinatorics and Probability questions are characterized by panels with numbers, often concluding with requests for the total number of possible combinations or probability estimations based on digit variations.\medskip

\begin{table}
\caption{\label{tab:mathematical_benchmarks}Mathematical benchmarks and illustrative examples.}
\begin{indented}
\item[]\begin{tabular}{@{}lp{0.33\linewidth}p{0.33\linewidth}}
\br
\textbf{Benchmark} & \textbf{Description} & \textbf{Example} \\
\mr
Geometry and Figures &
Calculation and exploration of spatial properties of shapes from triangles to cubes, examining geometric transformations and dimensional relationships &
\textit{In the figure, $PQ = QS = SP = ST$; $\angle STP = 42^\circ$; $QR$ is perpendicular to $RS$ and $RST$ is a straight line. What is the measure of angle $\angle PQR$?} \\

Visual Algebra and Arithmetic &
Transformation of mathematical equations into graphical representations, solving numerical problems through intuitive spatial reasoning &
\textit{We have three cards with four-digit numbers. As shown in the figure, three digits are covered. The sum of the three four-digit integers is 10126. What are the hidden digits?} \\

Visual Logic and Reasoning &
Interpretation and solution of complex puzzles through systematic deductive analysis of visual relationships and logical connections &
\textit{An ant moves from point $A$ to point $B$. It cannot move twice on the same segment. In how many ways can it reach from $A$ to $B$?} \\

Patterns and Sequences &
Identification and prediction of logical progressions in visual and numerical arrangements through systematic pattern recognition &
\textit{A spiral of consecutive numbers is created, as shown, starting from $1$. If the spiral pattern continues, in what form will the numbers $625$, $626$, and $627$ appear?} \\

Combinatorics and Probability &
Analysis of possible arrangements and event probabilities by exploring complex combinations under specific mathematical constraints &
\textit{Carmen labeled the vertices of a square-based pyramid with the numbers $1, 2, 3, 4$, and $5$, one for each vertex. For each face, she calculated the sum of the numbers on its vertices. Four of these sums are $7$, $8$, $9$, and $10$. What is the fifth sum?} \\
\br
\end{tabular}
\end{indented}
\end{table}

Whether purely textual or accompanied by images, each question was individually presented to the models in structured prompts designed to encourage explicit reasoning before selecting an answer. Regardless of whether models possess inherent reasoning capabilities, an explanation was always requested before their final response. Model outputs were subsequently compiled and compared against the ground truth to assess accuracy rigorously. It is important to highlight that all prompts were consistently presented in the same format and language as the evaluated question. Despite this multilingual evaluation, we underscore the continued predominance of English as the primary training language, raising questions about its influence on model performance across different languages.\medskip

We employed various computational resources tailored to each model's size and evaluation requirements. Larger LLMs were accessed via dedicated APIs on cloud-based platforms such as Google Cloud, Azure, Nvidia, and Alibaba Cloud, selected primarily due to the availability of specific models. Local computing resources for utilizing the Transformers library were employed for smaller models (under 30B parameters) to streamline evaluation and facilitate rapid iteration.\medskip

\section{Results}
\label{sec:results}
We present a detailed analysis of the evaluation results obtained from various experiments to assess the multimodal reasoning capabilities of several state-of-the-art vision-language models on Kangarooo questions. The results focus on key aspects: model performance across languages and difficulty levels, comparative analysis of model responses, visual versus textual performance, and precision by mathematical topic.\medskip

\begin{figure}
\centering
\includegraphics[width=1\textwidth]{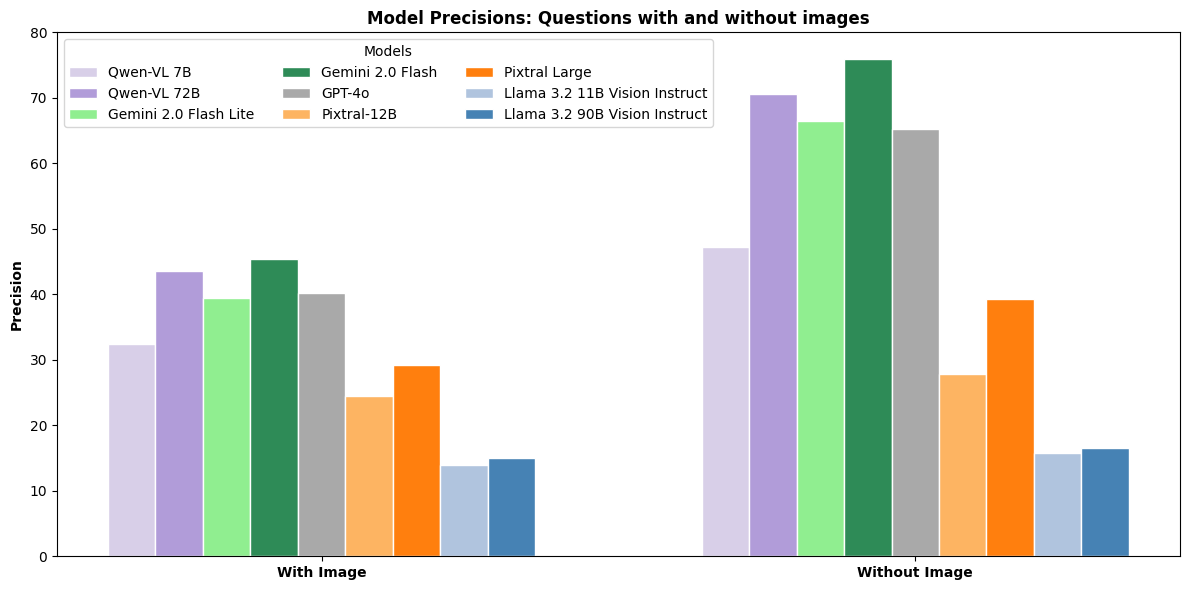}
\caption{\label{fig:with_image_vs_without_image}Comparison of model precision (\%) for questions with and without images. Bars represent the precision of each model in both conditions.}
\end{figure}

The analysis of various models’ performance in answering questions reveals notable differences in the interpretation of textual versus visual content. It is worth mentioning that we have considered three types of response: (1) correct answers supported by a reasoning, (2) wrong answers despite being accompanied by a reasoning, which results in being inaccurate, and (3) lack of answer.
First, we will analyze the models' accuracy in terms of correct answers: When the results are grouped, it is evident that models with more parameters, such as Qwen-VL 2.5 72B and Gemini 2.0 Flash, excel in handling text-based queries (70.6\% and 75.9\%, respectively). However, they exhibit significant declines when processing images (43.5\% and 45.4\%, respectively). In contrast, the Llama family, which overall presents a precision lower than random guessing, does not show significant improvement when considering only text-based questions (Llama 3.2 90B shows 15.0\% for image-based questions and 16.5\% for text-based questions). As shown in Figure~\ref{fig:with_image_vs_without_image}, overall, large-scale models such as the GPT, Gemini, and Qwen families demonstrate notable improvement on text-based questions, whereas their performance deteriorates on tasks requiring image analysis. In contrast, models with lower baseline performance show fewer differences between both modalities.\medskip

\begin{figure}
\centering
\includegraphics[width=0.8\textwidth]{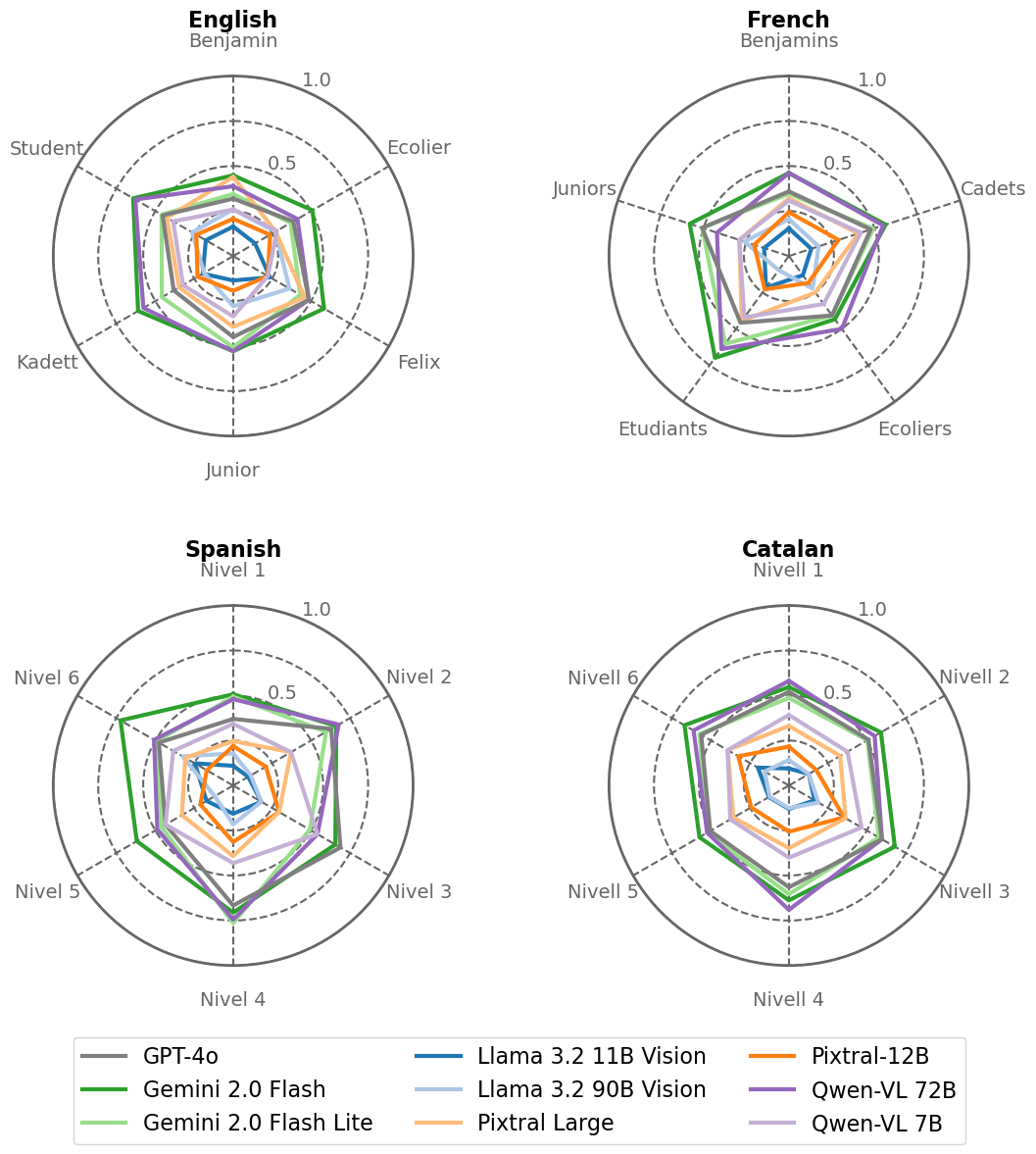}
\caption{\label{fig:radar_chart}Radar plots illustrating the performance of the multimodal LLMs on the visual Kangur evaluation across four languages (English, French, Spanish, and Catalan). Each radial axis corresponds to one of the designated difficulty levels (e.g.\ Benjamin, Cadets, Nivel~1, etc.).}
\end{figure}

Figure \ref{fig:radar_chart} presents charts comparing the performance of eight multimodal models across four languages. Despite they are not evaluated on the same questions in each language, this comparison highlights the disparities in performance depending on language and difficulty level, revealing distinct capabilities among the evaluated systems. Gemini 2.0 Flash demonstrates high consistency and robust results across all evaluated levels and languages, reflecting strong multilingual reasoning capabilities. \medskip

\begin{table}[t]
\centering
\caption{Accuracy (\%) of the different evaluated models across English, Spanish, French, and Catalan, with and without images. Left side of each cell shows image-based questions; right side shows non-image questions.}
\label{table}
\begin{tabular}{lllll}
\toprule
\textbf{Model} & \textbf{English} & \textbf{French} & \textbf{Spanish} & \textbf{Catalan} \\
\midrule
Qwen-VL 7B & 21.0 / 45.0 & 24.2 / 48.0 & 33.3 / 47.5 & 32.1 / 47.2 \\
Qwen-VL 72B & 36.6 / 71.1 & 35.4 / 70.1 & 40.0 / 68.2 & 44.9 / 71.4 \\
Gemini 2.0 Flash Lite & 32.0 / 58.7 & 35.4 / 60.5 & 43.0 / 61.9 & 37.9 / 68.0 \\
Gemini 2.0 Flash & 43.2 / 72.5 & 39.4 / 72.3 & 47.3 / 76.2 & 44.7 / 75.7 \\
GPT-4o & 30.5 / 55.7 & 25.3 / 65.0 & 40.6 / 62.3 & 40.1 / 66.3 \\
Pixtral-12B & 19.8 / 25.2 & 20.7 / 25.4 & 24.2 / 22.4 & 24.6 / 29.6 \\
Pixtral Large & 30.2 / 50.3 & 24.7 / 45.2 & 25.5 / 36.8 & 30.7 / 40.1 \\
Llama 3.2 11B Vision Instruct & 13.4 / 21.8 & 12.1 / 19.8 & 15.8 / 16.1 & 13.3 / 15.7 \\
Llama 3.2 90B Vision Instruct & 22.4 / 31.9 & 16.2 / 22.0 & 15.2 / 21.1 & 15.0 / 14.9 \\
\bottomrule
\end{tabular}
\end{table}

In addition to Gemini 2.0 Flash, the models with the highest average accuracy were Qwen-VL 72B and GPT-4o, achieving 54.6\% and 48.2\%, respectively, compared to 58.8\% for the leading model. Although Gemini 2.0 Flash performs best overall, there are specific scenarios where other models outperform it. For instance, Qwen-VL 72B achieves superior results at intermediate difficulty levels in certain languages, such as Level 4 in Spanish (74.5\% compared to 70.6\%) and the Ecoliers level in French (50\% compared to 43.2\%). Nonetheless,Gemini 2.0 Flash maintains a distinct advantage at both lower and higher difficulty levels, as seen in Level 6 in Catalan (67\% compared to 61.2\%) and the Ecoliers level in English (51\% compared to 41.2\%). \medskip

By contrast, smaller models such as Llama 3.2 11B Vision and Pixtral-12B exhibit notable limitations, particularly in addressing more complex questions. The lowest accuracies were recorded for Llama 3.2 11B (15.7\%), Llama 3.2 90B (20\%), and Pixtral-12B (23.7\%). Significant performance discrepancies also arise within model families. Pixtral Large outperforms Pixtral-12B (36.1\% compared to 23.74\%), Qwen-VL 72B significantly exceeds Qwen-VL 7B (54.6\% compared to 34.4\%), and Gemini 2.0 Flash surpasses its Lite variant (58.8\% compared to 47.7\%).\medskip

Language-specific variations reveal divergent behaviour among model families. The Gemini and Llama families show stronger results in Spanish than in Catalan: Gemini 2.0 Flash (62.9\% compared to 61.7\%), Gemini Flash Lite (53.8\% compared to 46.9\%), Llama 11B (15.9\% compared to 14.1\%), and Llama 90B (20.7\% compared to 14.4\%). Conversely, GPT and Qwen families demonstrate better performance in Catalan: GPT-4o (56.0\% compared to 50.6\%), Qwen 7B (40.1\% compared to 32.7\%), and Qwen 72B (59.3\% compared to 56.4\%). \medskip

An upward trend in accuracy is observed as the difficulty level increases. This phenomenon may be linked to the reduced presence of images in higher-level questions and a corresponding shift toward more abstract reasoning over visual representation in these contexts. \medskip

\begin{figure}
\centering
\includegraphics[width=0.9\textwidth]{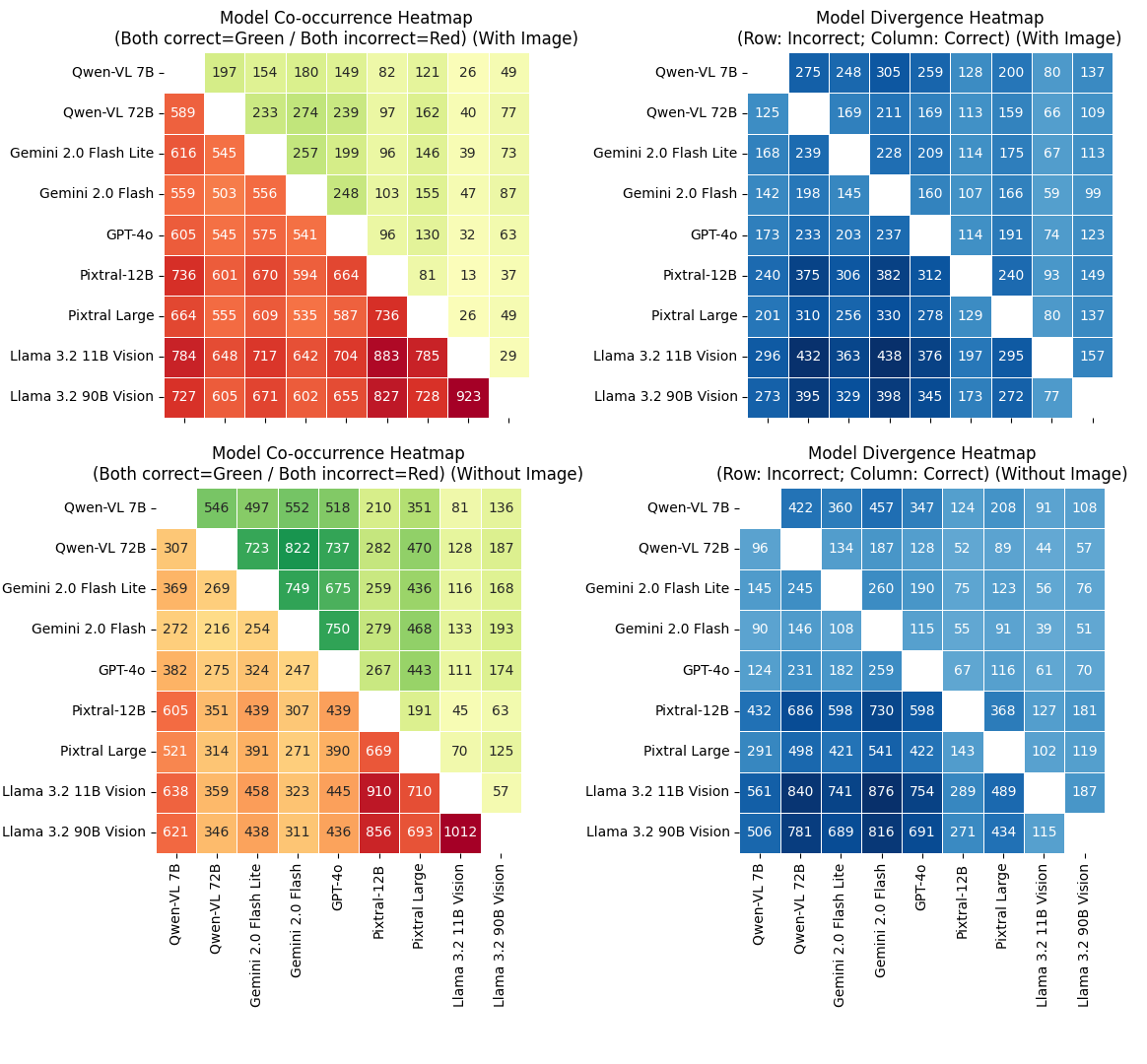}
\caption{\label{fig:model_vs_model_concurrence}Comparison of model co-occurrence heatmaps. \textbf{Left:} Co-occurrence of simultaneous success and failure. The upper triangle (greens) shows counts of instances where both models (row and column) answered the same question correctly. The lower triangle (reds) represents counts where the row model failed while the column model succeeded. \textbf{Right:} Frequency of advantage patterns. This heatmap visualizes how often a row model failed on a question that a column model answered correctly. Darker blue indicates more such occurrences, highlighting the relative advantage of the column model over the row model.}
\end{figure}

The co-occurrence heatmaps in Figure \ref{fig:model_vs_model_concurrence} provide a detailed comparative analysis of the models' performance, illustrating the alignment (left) or divergence (right) of model answers across identical problems. The figure should be interpreted separately with non-normalized values.\medskip

The left map is divided into an upper (green) triangular matrix and a lower (red) triangular matrix. The green one shows the frequencies where a pair of models simultaneously provided a correct answer for a question. At the same time, the red one counts the instances where a pair of models have provided an incorrect answer simultaneously. Unanswered questions are omitted from the analysis and are neither considered correct nor incorrect.
Notably, a higher concurrence rate in correct and incorrect responses is observed among models belonging to the same family but differing in scale, such as the Gemini and Qwen-VL variants, especially in the case of responses without an image. The concurrence rate is also higher for the highest-performing models. When Gemini 2.0 Flash succeeds, Qwen-VL 2.5 72B also succeeds in 42.3\% of the occasions, and they both fail simultaneously 29.1\% of the time overall.\medskip

The graph on the right is also divided into two triangular matrices. However, they both represent the same metric: they count the frequencies when the row model fails and the column model succeeds simultaneously. This map is critical to assess whether a combination of models would complement each other, although the results show no perfect combination to achieve this goal. The highest concurrences occur for low-performance models against high-performance models: take the example of any Llama version and the Qwen family. When comparing pairs of larger models in this matrix, the cases where one fails but the other succeeds are less significant than their simultaneous success or failure rate. After evaluating Gemini 2.0 Flash and Qwen-VL 2.5 72B again, Gemini fails while Qwen succeeds 13\% of the time, and Qwen fails while Gemini succeeds 15\% of the time. Therefore, these presented co-occurrence patterns imply a limitation in achieving performance improvement through model ensembles due to models' tendency, especially high-performing ones, to exhibit correlated success and failure. Separating the matrices depending on the presence of images in each question, we observe that the trend is very similar for both cases, with the key difference being the imbalanced correct answers for the cases of questions without images, as all models perform better. \medskip


\begin{figure}
\centering
\includegraphics[width=0.9\textwidth]{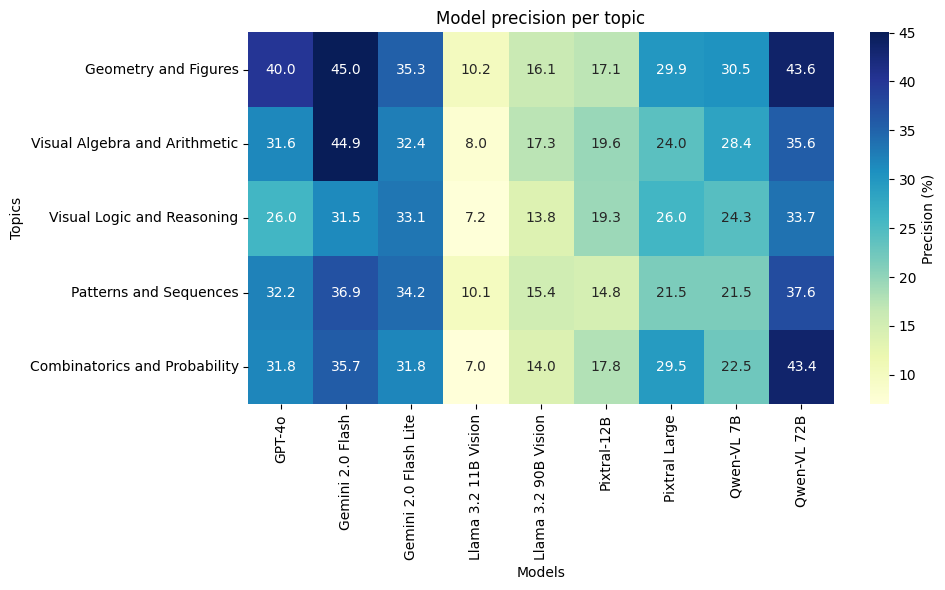}
\caption{\label{fig:model_precision_per_topic}Heatmap illustrating the precision (\%) of different models across various mathematical topics in questions with visual support. Topics are listed on the vertical axis, models on the horizontal axis, and the color gradient indicates the precision score.}
\end{figure}

In order to identify performance differences across specific mathematical domains on questions that include images, Figure \ref{fig:model_precision_per_topic} presents a heatmap of model precision indicating that no single model consistently achieves the highest precision in all areas. 
Qwen-VL 2.5 72B, demonstrates relatively stable performance, although it does not always lead in accuracy. Specifically, Qwen-VL 72B leads in Combinatorics and Probability with an accuracy of 43.4\%, and in Patterns and Sequences with 37.6\%. Conversely, Gemini 2.0 Flash outperforms other models in Geometry and Figures, achieving 45\% accuracy. Additionally, comparative analysis within the Qwen-VL model family reveals notable differences; Qwen-VL 72B scores 43.6\% in Geometry and Figures compared to 30.5\% by Qwen-VL 7B, while performance in Visual Logic and Reasoning declines from 33.7\% (Qwen-VL 72B) to 24.3\% (Qwen-VL 7B). Meanwhile, other models occasionally attain top precision but exhibit more significant variability depending on the specific task. This matrix may indicate that using different model families for different mathematical domains is impossible. Remarkably, the \textit{Visual Logic and Reasoning} category exhibits the lowest precision across all evaluated models, emphasizing its inherent complexity and the significant challenges current multimodal models face in solving visually presented logical reasoning tasks.\medskip

\section{Discussions}
\label{sec:discussions}
We have designed a benchmark test to evaluate the capabilities of MLLMs in solving mathematical problems across multiple languages. Our findings demonstrate that most models performed above chance-level accuracy, exhibiting moderate—and in some cases notably high—mathematical reasoning abilities in a multilingual context.\medskip

Nevertheless, overall performance remained moderate across evaluated areas such as geometry and visual algebra, with no single model excelling uniformly across all topics. Particularly noteworthy is the limited improvement observed when models handled purely textual problems —even those presented as images with text and mathematical symbols— compared to those involving visual inputs, suggesting that these models often underutilized diagrammatic and visual information.\medskip

We have also identified substantial variability in performance depending on language and problem difficulty. While most models effectively solved simple problems, they faced notable challenges when interpreting figures demanding advanced spatial visualization or comprehending three-dimensional representations. Among evaluated models, Gemini 2.0 Flash demonstrated the highest accuracy in image-based tasks, followed closely by Qwen-VL 2.5 72B and GPT-4o, though none reached human-level precision.\medskip

A comparison between proprietary and open-source models reveals trade-offs. Proprietary models’ rapid API integration and large datasets yield high textual precision, though visual interpretation remains challenging. Open-source models offer greater customization and balanced multimodal results but demand more technical resources and maintenance.\medskip

This study has shown notable shortcomings in MLLMs concerning their capacity to solve mathematical problems that include visual elements. Although these models can extract basic information such as text and numbers from images, they encounter substantial difficulties interpreting three-dimensional figures and reasoning about complex spatial relationships. While models correctly answer questions where figures serve merely as complementary visual support and essential information resides in the text, their performance notably declines in tasks requiring precise measurements derived from scales or grids explicitly represented within the figures. Specific limitations are evident in recognizing complementary angles in two-dimensional geometric figures derived from given angles.\medskip

Further challenges arise when models confront shapes involving curvature or angles distinct from right angles, reflecting a dependency on simple, rectilinear figures such as squares as primary reference units. These models frequently miscalculate areas resulting from overlapping figures due to an inadequate perception of magnitudes and essential proportional relationships. Precision notably decreases with increasing complexity and detail in the figures, yet accuracy remains limited even in simpler figures with fewer elements. In two-dimensional contexts, although models recognize points as intersections of lines, they fail to fully comprehend figures formed by multiple intersections. The stacking or overlapping of elements, which requires spatial visualization, is consistently misunderstood, with all elements incorrectly perceived as existing in a single two-dimensional plane. These outcomes highlight a superficial and limited understanding of visual and mathematical representations, emphasizing that substantial research and development efforts remain necessary to achieve a more profound and more accurate comprehension of mathematical problems and their associated visual representations.\medskip

To assess whether models genuinely engage in reasoning rather than relying on rote memorization, several systems were evaluated using the Kangaroo tests from the Comunitat Valenciana, conducted during the first week of April. This offered an opportunity to observe model behavior on unfamiliar and nuanced problems. Pixtral and Llama, for instance, frequently returned "No answer" responses. These models often attempt a reasoning process, but when the derived solution does not correspond to any of the provided options, they default to "No answer". In contrast, when they do not engage in reasoning, they tend to select randomly from the available choices. Meanwhile, models such as Gemini and GPT-4o demonstrated more coherent and structured reasoning, enabling them to consistently arrive at the correct answers. Their performance aligns closely with results from previous years' datasets, suggesting that their success is grounded in genuine reasoning rather than pattern memorization.

\section*{Data Availability Statement}

The dataset and codes supporting this study are available at \href{https://github.com/mc-flai/kangur-ai-evaluation.git}{Kangaroo Language Test on GitHub}.

\section*{Conflict of interest}
The authors declare no conflict of interest.

\section*{Acknowledgements}

We thank the Kangaroo Committee of the Societat Catalana de Matemàtiques (SCM) and, especially, the ``Comissió Cangur del País Valencià'', which is part of the former, for the work they do in preparing the test each year and for the useful discussions we have had regarding certain aspects related to the statements of this test.

\section*{Funding}
AIS acknowledges the financial support from the European Union, Next GenerationEU, under the public subsidies of the ``Programa Investigo'', within the framework of the Recovery, Transformation, and Resilience Plan (Reference INV/2023/25). SH acknowledges the funding of project PID2021-128676OB-I00 and JAC acknowledge funding of project PID2022- 138860NB-I00  by FEDER/MCIN, funded by MCIN/AEI/10.13039/ 
501100011033 and by ``ERDF A Way of Making Europe'' by the European Union. RV the funding provided by Digital Futures, in their Demonstrator-project program. MJFA acknowledges the financial 
support of the Generalitat Valenciana Project grant
CIAICO/2022/252. Funding for open access charge: CRUE-Universitat Politècnica de València

\section*{References}

\end{document}